\documentclass[9pt,conference,a4paper]{IEEEtran}
\IEEEoverridecommandlockouts
\usepackage{cite}
\usepackage{amsmath,amssymb,amsfonts}
\usepackage{algorithmic}
\usepackage{graphicx}
\usepackage{textcomp}
\usepackage{xcolor}
\usepackage{bbm}
\usepackage{algorithm, algorithmic}
\usepackage{stfloats}
\usepackage[square, comma, sort&compress, numbers]{natbib}
\usepackage{subfigure}
\usepackage{makecell}
\usepackage{multirow}
\def\BibTeX{{\rm B\kern-.05em{\sc i\kern-.025em b}\kern-.08em
    T\kern-.1667em\lower.7ex\hbox{E}\kern-.125emX}}
\begin{document}

\title{
Point Cloud Attribute Compression via Successive Subspace Graph Transform
\thanks{978-1-7281-8068-7/20/\$31.00 \copyright2020 IEEE}
}

\author{\IEEEauthorblockN{Yueru Chen\IEEEauthorrefmark{1},
Yiting Shao\IEEEauthorrefmark{2}, Jing Wang\IEEEauthorrefmark{1},
Ge Li\IEEEauthorrefmark{2},
and C.-C. Jay Kuo\IEEEauthorrefmark{3}}
\IEEEauthorblockA{\IEEEauthorrefmark{1} Peng Cheng Lab, Shenzhen, China}
\IEEEauthorblockA{\IEEEauthorrefmark{2}{Peking University, Shenzhen Graduate School, China}}
\IEEEauthorblockA{\IEEEauthorrefmark{3}{University of Southern California, Los Angeles, California, 
USA}}}


\maketitle

\begin{abstract}

Inspired by the recently proposed successive subspace learning (SSL) principles, we develop a successive subspace graph transform (SSGT) to address point cloud attribute compression in this work. The octree geometry structure is utilized to partition the point cloud, where every node of the octree represents a point cloud subspace with a certain spatial size. We design a weighted graph with self-loop to describe the subspace and define a graph Fourier transform based on the normalized graph Laplacian. The transforms are applied to large point clouds from the leaf nodes to the root node of the octree recursively, while the represented subspace is expanded from the smallest one to the whole point cloud successively. 
It is shown by experimental results that the proposed SSGT method offers better R-D performances than the previous Region Adaptive Haar Transform (RAHT) method.

\end{abstract}


\section{Introduction}

Three dimensional (3D) object processing has received a lot of attention in recent years due to the rapid development in 3D autonomous driving, gaming, and remote sensing. The point cloud model is one of the most popular data formats to represent a 3D object in many applications, because of its easy access and complete description in the 3D space. Point cloud models are always composed of a large number of points associated with attributes (\emph{e.g.}, color), leading to huge storage space and high computational cost. It makes the compression of point clouds to be a critical and valuable research topic.
 
Point cloud compression has been widely studied in the research community. The octree structure is an effective representation of point clouds and has been adopted in point cloud geometry coding, \emph{e.g.} \cite{peng2005technologies}\cite{schnabel2006octree}.  In this paper, we use this well-developed octree method for geometry coding. 
As for attribute compression, many efforts have been made to explore the relationships among points based on the octree structure. Zhang \emph{et al.} \cite{zhang2014point} designed graphs at a certain level of the octree, which uses the subset of points as the vertex set and connects points within the threshold distance. Then graph transform is applied to encode point cloud attributes. The transform scheme provided better performance over traditional discrete cosine transform (DCT) in point cloud compression. Their way to construct the graph would create isolated sub-graphs when point clouds are sparse. To address this problem, Cohen \emph{et al.} \cite{cohen2016attribute} proposed a K-nearest neighbor (KNN) method by adding edges between more distant points in the graph.

For those one-stage graph transform methods, better performance is reported when constructing a larger graph at the higher level of the octree, while the higher computational cost is required simultaneously. 
The region adaptive hierarchical transform (RAHT) \cite{de2016compression} is proposed by Ricardo \emph{et al.} to achieve the more efficient attribute compression via the multi-stage transform scheme based on octree structure. However, the transforms simply apply on two nodes in each stage, which is weak to fully represent the spatial correlations among points.

Motivated by the successive subspace learning (SSL) framework in the subspace learning field \cite{chen2020pixelhop}, we design a new hierarchical transform method, called successive subspace graph transform (SSGT) for point cloud attribute compression. Existing works \cite{kuo2019interpretable,zhang2020pointhop} using SSL have been proposed to solve image and point cloud classifications. The set of effective features of successively growing subspace points or patches are extracted through multi-stage Saab (subspace approximation with adjusted bias) transform \cite{kuo2019interpretable} and utilized to address the classification problems. Principled by SSL, the proposed SSGT takes advantage of the “successive subspace growing" process to handle large point clouds. The graph of a larger subspace is constructed on those of its constituent subspace of smaller sizes, offering an effective description of relationships among points.

These are two major contributions of this work. First, we propose a novel way to construct a weighted graph with self-loop to represent the subspace of the point clouds and the graph Fourier transform defined on the normalized Laplacian is utilized to encode the color attributes.
Second, we adopt the SSL framework for point cloud attribute compression and introduce the successive subspace graph transform scheme. It provides a more efficient and effective way to represent the spatial correlations among a large number of points.

The rest of the paper is organized as follows. The proposed successive subspace graph transform (SSGT) is presented in Sec.\ref{sec2}. A comparison between SSGT and RAHT is discussed in Sec. \ref{raht}. Experimental results are shown in Sec. \ref{sec3}.  Concluding remarks and future works are given in Sec.\ref{sec4}. 

\section{Proposed SSGT method}\label{sec2}
\begin{figure*}[tb]
\centerline{\includegraphics[width=0.85\linewidth]{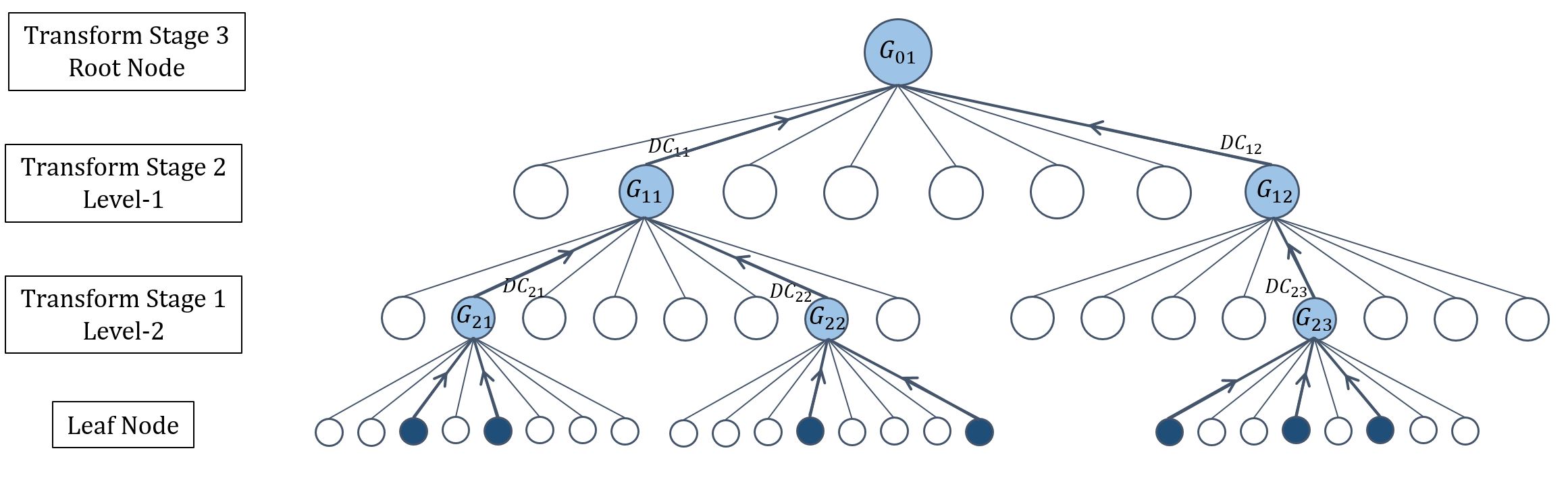}}
\caption{The diagram of the proposed SSGT framework.}\label{fig:overview}
\end{figure*}
\subsection{Framework overview}
The framework overview of the proposed SSGT method is shown in Fig. \ref{fig:overview}. To transform a point cloud model consists of a large number of points, we first apply the octree decomposition, and the point cloud decomposes into eight octants recursively to form an octree-tree structure with its root being the whole point cloud and its occupied leaf being a point. Those occupied nodes at different levels represent a subspace of the point cloud model with different spatial sizes. We design a weighted graph with self-loop to describe the subspace and define the graph Fourier transform based on the symmetric normalized Laplacian. The attribute signal (\emph{e.g.}, color) can be transformed to the graph spectral domain with the energy compaction property. The ﬁrst-stage graph transform is applied at the bottom of the octree. Then, multi-stage transforms are processed from all child nodes to their parents stage by stage until the root node is reached. The proposed SSGT framework successively learns the point cloud subspace and offers an efficient and effective way for point cloud compression.

\subsection{Graph Construction} \label{graph_construction}
{\bf Problem Formulation.}
We define a point cloud of $N$ points as ${\bf P}=\{{\bf p}_1,..,{\bf p}_N\}$, where ${\bf p}_n \in \mathbb{R}^{3}$. Assume ${\bf P}$ occupies the space of size $1 \times 1 \times 1$, and can be partitioned using the octree decomposition with depth $L$. The octree will contain $2^{3L}$ leaf nodes with $N$ occupied ones and the spatial size of the leaf node is $2^{-L}\times2^{-L}\times2^{-L}$. 

At the level $l$, the $k$th node represents a subspace of the point cloud of size $2^{-l}\times2^{-l}\times2^{-l}$. we can construct a weighted graph with self-loop,  $\mathcal{G}_{lk}$, to describe it, where the vertex set is composed of all occupied child nodes. The designed graph model the relationships among the nodes and the corresponding graph Fourier transform, $\Phi_{lk}$, is calculated based on the symmetric normalized graph laplacian.
The multi-stage transforms can be applied from level $(L-1)$ to the root level, the represented subspace expands from size $2^{-(L-1)}\times2^{-(L-1)}\times2^{-(L-1)}$ to $1 \times 1 \times 1$ accordingly.
\begin{figure}[tb]
\centerline{\includegraphics[width=0.65\linewidth]{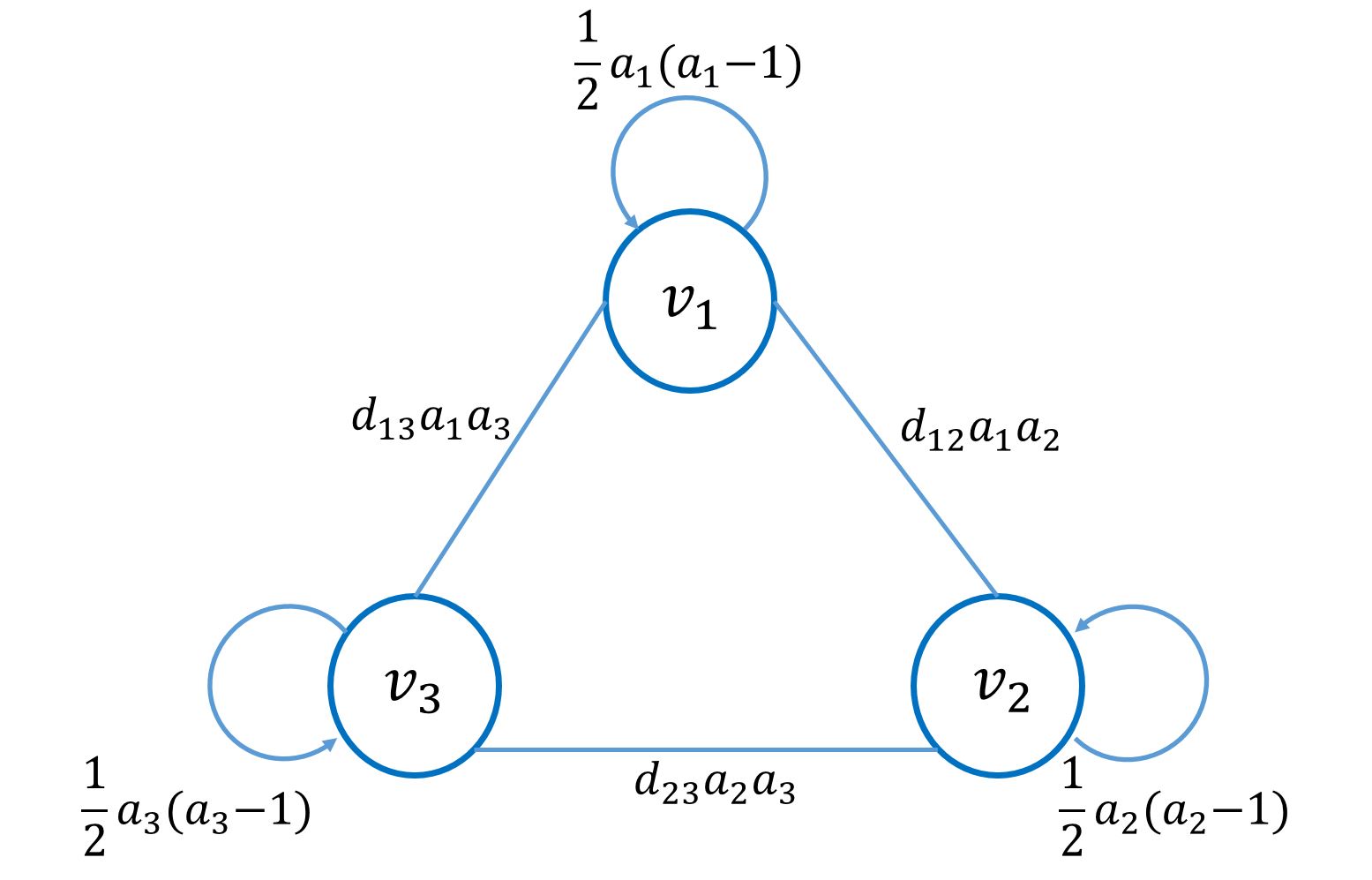}}
\caption{Illustration of the graph construction.}\label{fig:graph}
\end{figure}

{\bf Designed Graph.}
The graph design is illustrated in Fig. \ref{fig:graph}. For the $k$th node at the $l$th level, we can construct a weighted graph with self-loop, $\mathcal{G}_{lk}=\{\mathcal{V}_{lk},\mathcal{E}_{lk},W_{lk}\}$, where $\mathcal{V}_{lk}=\{v_0,...,v_{N_{lk}}\}$ denotes a set of vertices, which contains all occupied child nodes at level $(l+1)$. The set of edges is defined by
$\mathcal{E}_{lk} = \{ (v_i,v_j) \vert (v_i,v_j) \in \mathcal{V}_{lk}^2 \wedge  v_i \neq v_j \}$. $W_{lk} \in \mathbb{R}^{N_{lk} \times N_{lk}}$ denotes a edge weight matrix and $N_{lk}$ denotes the number of vertices. 

The spatial position of the $v_i$ is denoted as ${\bf x}_i$ and the number of the points in the $v_i$ is denoted as $a_i$. The weight matrix $W_{lk}$ is defined by
\begin{equation}\label{eq:weight}
\begin{aligned}
W_{lk}(i,j) = \left\{ \begin{array}{rcl}
{1 \over 2} \times a_i\times (a_i-1) & \mbox{for} & i=j \\
d_{ij} \times a_i \times a_j & \mbox{for} & i \neq j
\end{array}\right.,
\end{aligned}
\end{equation} 
\begin{equation}\label{eq:distance}
\begin{aligned}
d_{ij} = e^{-\alpha \| {\bf x}_i - {\bf x}_j \|^2}.
\end{aligned}
\end{equation}

Considering the relationships among vertices, the edge weights are designed in two aspects: 1) the number of points (\emph{i.e.}, $a_i$) within each vertex, and 2) the distances between vertex pairs. 
In the first aspect, we assume each point in the subspace have relations with all other points, representing by the sub-edges to connect every point pairs. In $i$th vertex, there are ${1 \over 2}\times a_i\times(a_i-1)$ sub-edges among points. From $i$th vertex to $j$th vertex, there are $a_i\times a_j$ sub-edges among points. The number of sub-edges is introduced to the Eq. \ref{eq:weight} to represent the impact of different numbers of points in vertices.
In the second aspect, distances between vertex pairs are utilized as expressed in Eq. \ref{eq:distance}, where the parameter $\alpha$ can control the speed of the edge weight decay. And the same $\alpha$ is adopted over all stages of transforms. In the experiments, we simply use the voxelized position of the vertices, and the voxel grid size increases from $2^{-L}$ to $2^{-1}$ when applying the transform from the level $(L-1)$ to the root node recursively. 

{\bf Graph Fourier transforms.}
Based on the designed graph, we can compute the degree matrix $D_{lk}= diag(d_1,..., d_{N_{li}})$, where $d_i= a_i (a_i-1) + \sum_{j\neq i} d_{ij} a_i a_j$. The symmetric normalized Laplacian matrix \cite{chung1997spectral} is expressed as
\begin{equation}
\begin{aligned}
L_{lk}(i,j) = \left\{ \begin{array}{rcl}
D_{lk}(i,i) - 2 \times W_{lk}(i,i)  & \mbox{for} & i=j \\
-W_{lk}(i,j) & \mbox{for} & i\neq j \\
\end{array}\right.,
\end{aligned}
\end{equation} 
\begin{equation}
L^{sym}_{lk} = D_{lk}^{-{1 \over 2}}L_{lk}D^{-{1 \over 2}}_{lk}.
\end{equation}

The eigenvalue decompoistion of $L^{sym}_{lk}$:
\begin{equation}
L^{sym}_{lk} = \Phi_{lk}  \Lambda_{lk} \Phi ^{{T}}_{lk},
\end{equation}
where $\Phi_{lk} = \{{\bf t_0},...,{\bf t_{N_i}}\}$ is the matrix whose column is the eigenvector of $L^{sym}_{lk}$ and $\Lambda_{lk}  = diag( \lambda_i)$ is the matrix whose diagonal terms are the corresponding eigenvalues and are sorted in an increasing order. The graph signal can be transformed by $\Phi$ into the graph spectral domain.

The reason for using the normalized Laplacian matrix is to introduce the effect of the self-edges in the graph. Based on the properties of the symmetric normalized Laplacian matrix \cite{von2007tutorial}, we have $\lambda_0 = 0$ and its eigenvector ${{\bf t_0} = D^{1 \over 2} \mathbbm{1}}$. We can treat ${\bf t_0}$ as the DC transform kernel, and the DC coefficient is the weighted average of the input graph signal. The other eigenvectors as the AC 
transform kernels to describe the high frequency details. In the proposed SSGT method, only the DC coefficients are propagated to the next stage.

\subsection{Multi-stage transforms} \label{subsec:multi-stage}
To transform large point clouds, we use the multiple subspace graph transforms in cascade to expand the subspace size gradually. The subspace transform starts at level $(L-1)$ of the octree and applied to their occupied child node in the $L$th level which is a point. The attribute signal defined on the point can be transformed accordingly. The DC coefficients are fed into the upper level as a new signal defined on the nodes at level $(L-1)$. This process can be repeated stage by stage until the root node is reached.

The speed of the subspace expansions can be adjusted in the proposed framework. If we learn the graph transforms at every level of the octree, the size of the subspace represented by the parent nodes is eight times larger than that of their child nodes. We can increase the expansion speed by applying the subspace transform for every two levels, which leads to eight times faster than the previous settings. However, it should be noticed that if the expansion speed is too fast, the represented graph will contain too many vertices, and its computational cost will increase. In contrast, the capability of the small graph is limited to model the subspace well and it would provide the worse compression performance. 

\subsection{Attribute Compression}

The point cloud attributes can be treated as the graph signal defined on points. In a subspace, we can define the signal $F_{lk}$ on the constructed graph $\mathcal{G}_{lk}$ as $f: V_{lk} \rightarrow R, f_i=f(v_i)$. The signal $F_{lk}$ can be transformed to $H_{lk} =\Phi_{lk}^T F_{lk}, H_{lk} = \{h_0,...,h_{N_{lk}}\}$. $h_0$ is the DC coefﬁcient, and $h_1,...h_{N_{lk}}$ are all AC coefficients. Only $h_0$ will be propagated to the upper level as a new graph signal.

For example, when encoding the color information of point clouds, at level $(L-1)$, $F_{L-1,k}$ can be the color values (\emph{e.g.}, luminance values) at points in the subspace represented by the $k$th node. Then the DC coefficients of $F_{L-1,k}$ would be a new signal defined on the $k$th node and will be further encoded in the next stage. Finally, all AC coefficients and a DC coefficient of the root node are quantized and entropy-coded.

\begin{figure*}[ht!]
\begin{minipage}[b]{0.45\linewidth}
  \centering
  \centerline{\includegraphics[width=0.85\linewidth]{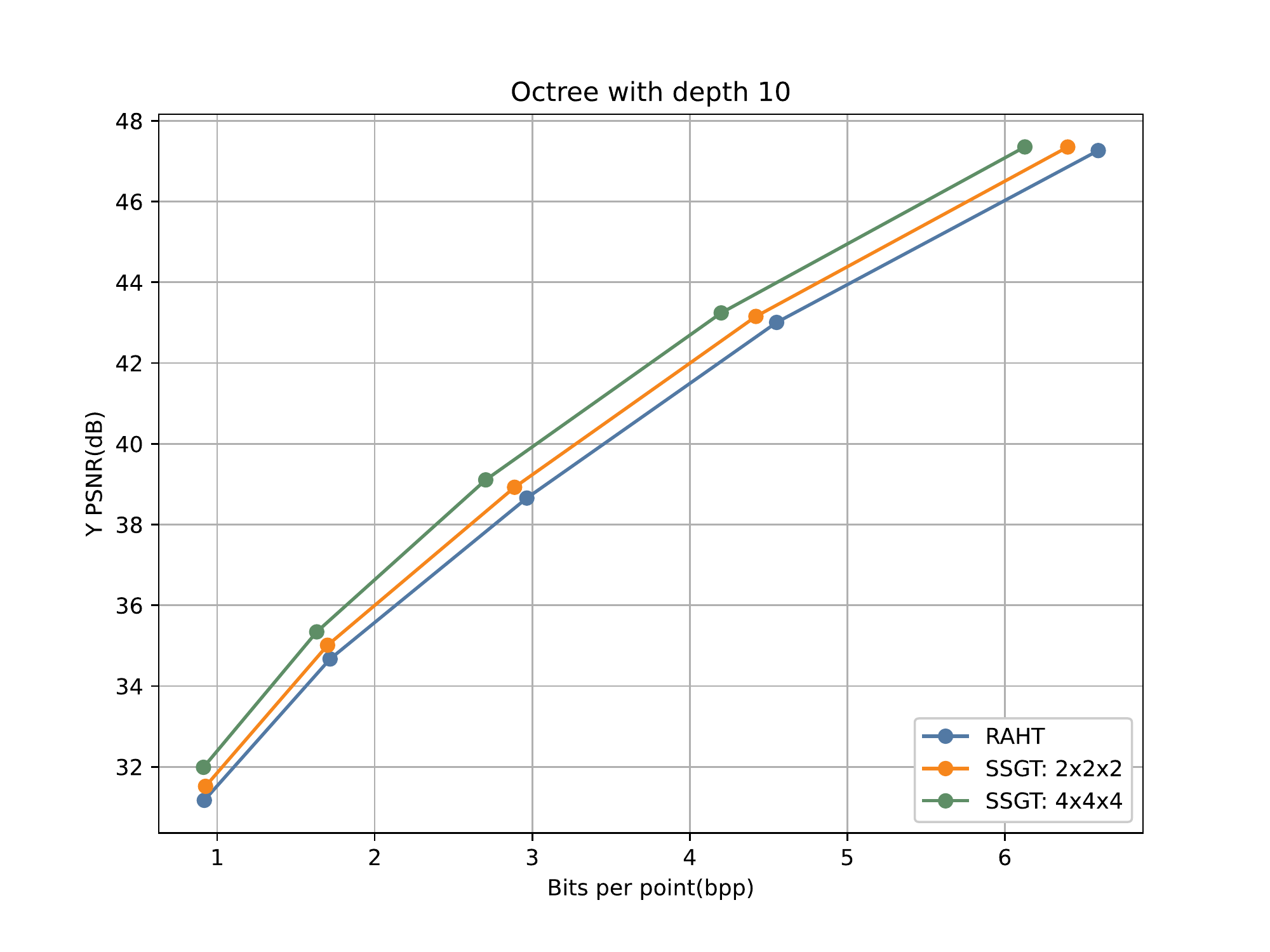}}
\end{minipage} 
\hfill
\begin{minipage}[b]{0.45\linewidth}
  \centering
  \centerline{\includegraphics[width=0.85\linewidth]{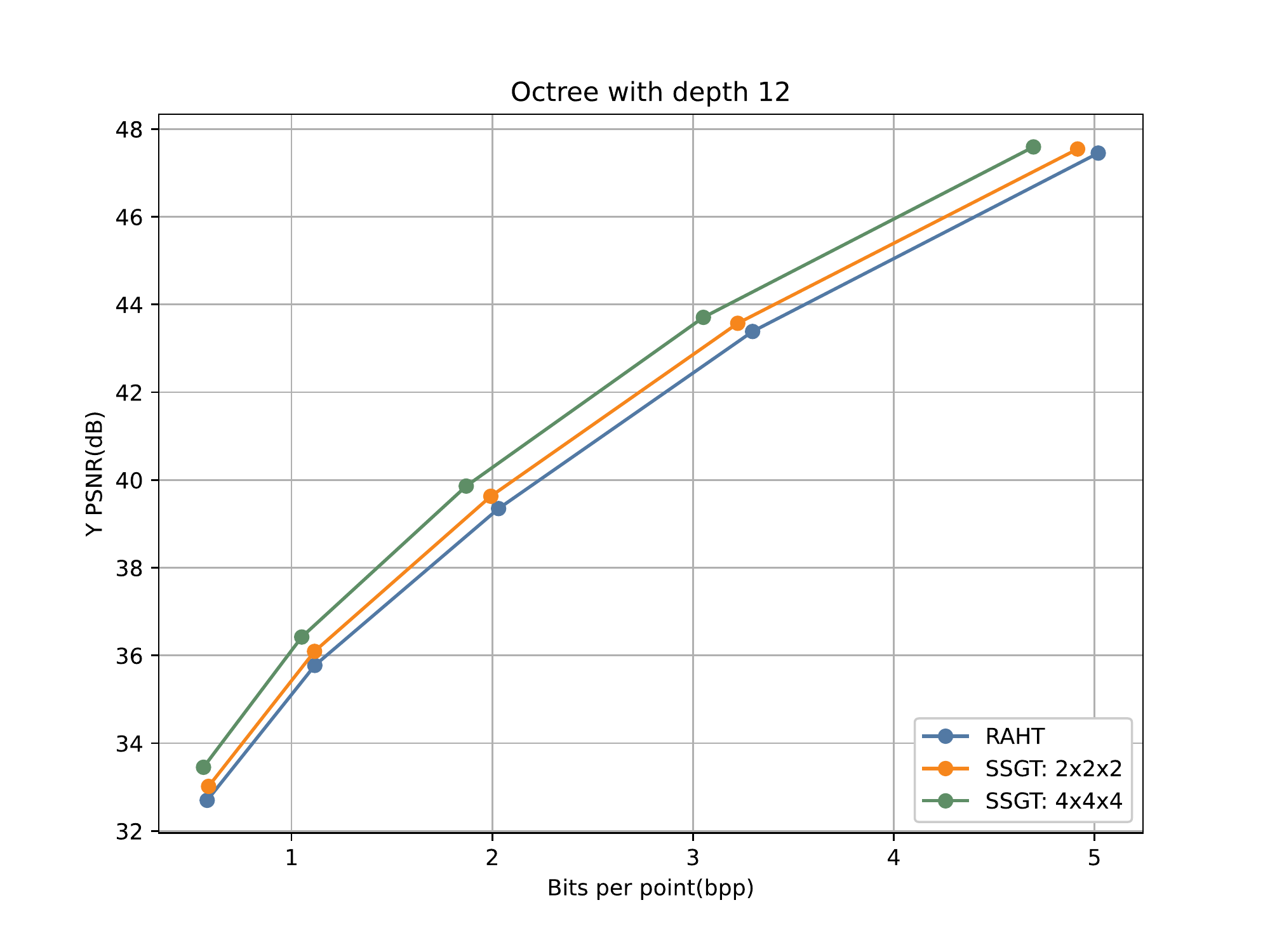}}
\end{minipage}
\caption {Compression performance (in dB) vs. bitrate (in bpp) on the "longdress" frame with five different quantization steps, where the blue line indicates the RAHT approach and the orange and green lines indicate the SSGT method with subspace size $2\times2\times2$ and $4\times4\times4$ voxels, respectively.}
\label{fig:curves}
\end{figure*}
\subsection{Comparison with RAHT}\label{raht}
RAHT is a special case of the proposed SSGT scheme. In every level of the octree, the subspace is further partitioned in x,y,z directions, and the created graphs only contain two vertices, denoted as $v_1$ and $v_2$. The corresponding degree matrix, Laplacian matrix, and the normalized Laplacian matrix are expressed as following:
\begin{equation}
{L=\begin{pmatrix}
a_1a_2  & - a_1a_2\\
- a_1a_2 & a_1a_2
\end{pmatrix}},    
\end{equation}
\begin{equation}
 {D=({a_1+a_2-1}){\begin{pmatrix}
a_1  &0\\
0 & a_2
\end{pmatrix}}},   
\end{equation}
\begin{equation}
  L^{sym}={a_1a_2\over{a_1+a_2-1}}{\begin{pmatrix}
1\over a_1  &-{1\over\sqrt{a_1a_2}}\\
-{1\over\sqrt{a_1a_2}} & 1\over a_2
\end{pmatrix}},  
\end{equation}
where $a_1$ and $a_2$ represent the number of points in $v_1$ and $v_2$ respectively.
Calculate the eigenvalue decomposition of $L^{sym}$, the graph transform matrix $\Phi$ and the eigenvalue matrix is
\begin{equation}
{\Phi={1 \over \sqrt{a_1+a_2}}\begin{pmatrix}
\sqrt a_1  & - \sqrt a_2\\
 \sqrt a_2 & \sqrt a_1
\end{pmatrix}},    
\end{equation}
\begin{equation}
{\Lambda={\begin{pmatrix}
0  &0\\
0 & {a_1+a_2}\over{a_1+a_2-1}
\end{pmatrix}}}.
\end{equation}

$\Phi$ is the same as the transform matrix in RAHT. There are two limitations of only adopting two-vertices transforms in RAHT: 1) it treats the x, y, z directions differently, which causes the performance difference when changing the order of the transforms along x, y, and z axes\cite{zhang20193d}, and 2) this simple graph neglects the influence of the pairwise distances among points leading to weak modeling of the point relationships in the subspace.

\begin{figure*}[htb]
\centerline{\includegraphics[width=0.6\linewidth]{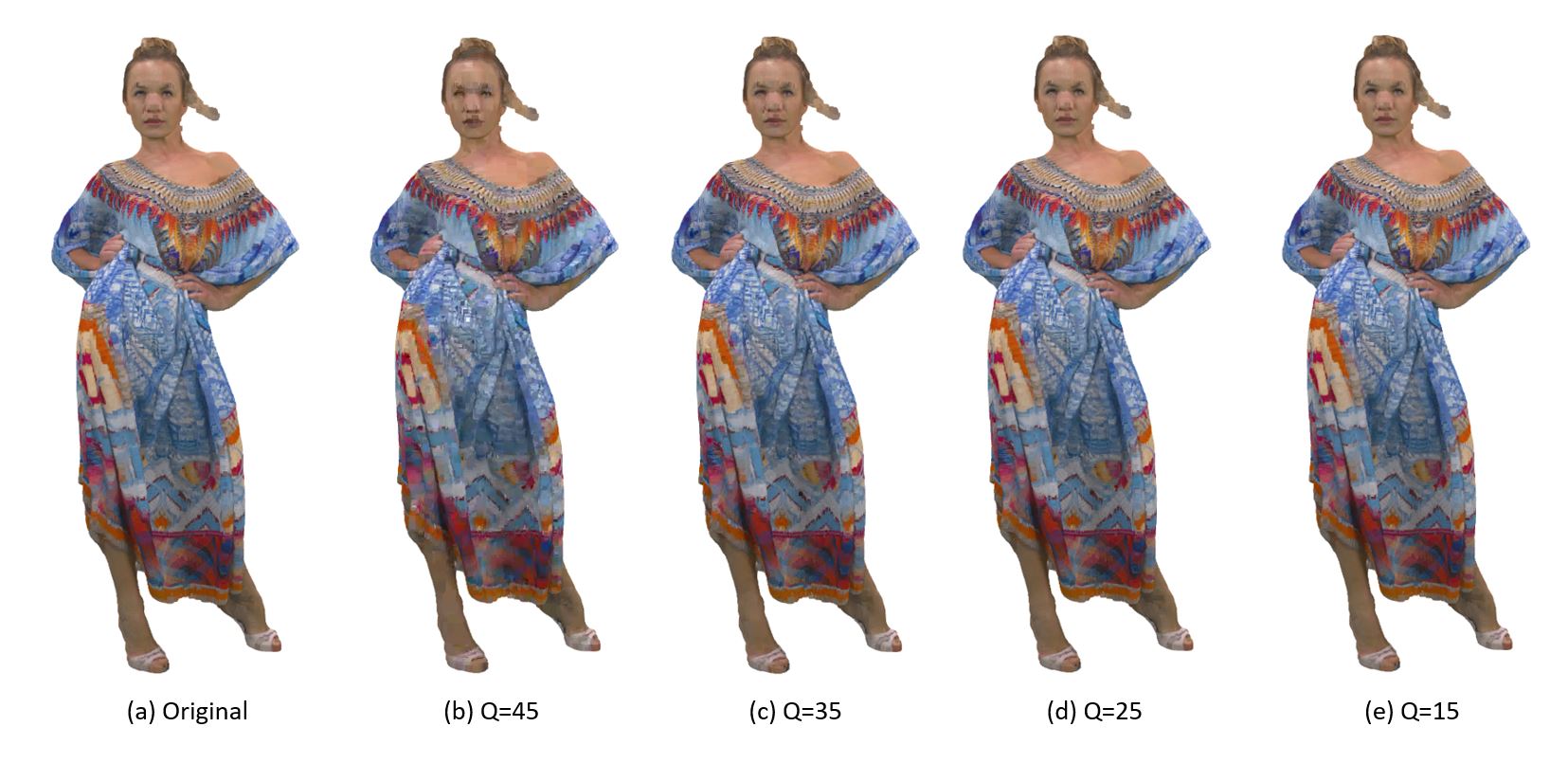}}
\caption{ Rendering results of the SSGT method with different quantization steps, where Q is the quantization step size. The subspace is partitioned into $4\times4\times4$ voxels in each stage transform.  
}\label{fig:rendering}
\end{figure*}

\section{Experiments}\label{sec3}
We test the proposed SSGT method for the color attributes compression using frames extracted from four sequences: “longdress”,  “redandblack”, “loot” and “soldier”, in 8i dynamic point cloud datasets \cite{d20178i}. Each point cloud model is voxelized by two different resolutions and decomposed as the octree with depth 10 and 12 respectively. The color attributes are firstly converted to YCbCr color space, and each color channel is processed independently. We adopt the R-D performance to evaluate the compression algorithms, where the compression rate is reported in bits per point (bpp) when encoding the Y, Cb, Cr channels, and the distortion is reported in peak signal-to-noise ratio (PSNR) of the luminance Y. Fig.\ref{fig:rendering} shows the rendering results of using the proposed SSGT method on the "longdress" frame with different quantization steps. 

\begin{table}[ht]
\caption{Comparison of R-D performance on point clouds with the octree of depth 10.}\label{tb:10}
\begin{center}
\begin{tabular}{|l|c|c|c|c|c|}
    \hline
    {\multirow{2}{*}{\textbf{Test Data}}} & \multicolumn{2}{|c|}{\textbf{SSGT}} & \multicolumn{2}{|c|}{\textbf{RAHT}} \\
    \cline{2-5}
    & \makecell{ Y-PSNR \\ (dB)} & \makecell{Bitrate \\ (bpp)} & \makecell{Y-PSNR \\ (dB)} &\makecell{ Bitrate \\ (bpp)}\\
    \cline{1-5}
    \text{ loot\_vox10} & 38.92 & 0.42  & 37.96 & 0.42 \\
    \cline{1-5}
    \text{redandblack\_vox10} & 38.50 & 0.80  & 37.74 & 0.88 \\
    \cline{1-5}
    \text{ soldier\_vox10} & 37.28 & 0.69  & 36.07 & 0.70 \\
    \cline{1-5} 
    \text{longdress\_vox10} & 35.34 & 1.63  & 34.67 & 1.72 \\
    \cline{1-5} \hline
\end{tabular}
\end{center}
\end{table}
\begin{table}[ht]
\caption{Comparison of R-D performance on point clouds with the octree of depth 12.}\label{tb:12}
\begin{center}
\begin{tabular}{|l|c|c|c|c|c|}
    \hline
    {\multirow{2}{*}{\textbf{Test Data}}} & \multicolumn{2}{|c|}{\textbf{SSGT}} & \multicolumn{2}{|c|}{\textbf{RAHT}} \\
    \cline{2-5}
    & \makecell{ Y-PSNR \\ (dB)} & \makecell{Bitrate \\ (bpp)} & \makecell{Y-PSNR \\ (dB)} &\makecell{ Bitrate \\ (bpp)}\\
    \cline{1-5}
    \text{ loot\_vox10} & 41.02 & 0.23 & 40.09 & 0.23 \\
    \cline{1-5}
    \text{redandblack\_vox10} & 39.95 & 0.50  & 39.30 & 0.54 \\
    \cline{1-5}
    \text{ soldier\_vox10} & 38.54 & 0.43  & 37.67 & 0.44 \\
    \cline{1-5} \hline
    \text{longdress\_vox10} & 36.42 & 1.05  & 35.77 & 1.12 \\
    \cline{1-5} \hline
    \end{tabular}
\end{center}
\end{table}
Table \ref{tb:10} and Table \ref{tb:12} show the R-D performance comparison of the proposed SSGT and RAHT method when the distortion is similar. For the proposed SSGT method, we apply the graph transforms at every two levels of the octree, where each corresponding subspace is partitioned into $4\times4\times4$ voxels. Generally, our proposed SSGT method provides consistent performance gains on four testing frames comparing with the RAHT method. In Fig. \ref{fig:curves}, we report the comparison of the R-D performance on the "longdress" frame by using five different quantization steps size, 15, 20, 25, 30, and 35. From the results, we can observe that the proposed SSGT method significantly outperforms RAHT at all quantization steps. Also, we compare two different subspace expansion speeds. The parent subspace is 8 times or 64 times the size of its child subspace, where the voxelized subspace contains $2\times2\times2$ voxels or $4\times4\times4$ voxels respectively. The results demonstrate that the larger constructed graph offers a better compression performance and description of the points relationships as discussed in Sec. \ref{subsec:multi-stage}.

\section{Conclusion and Future work}\label{sec4}
An attribute point cloud compression method based on the successive subspace graph transform (SSGT) was proposed in this paper. Experimental results demonstrate that the proposed method provides better attribute compression performance than the RAHT method in terms of compression rate and distortion. Moreover, the proposed method has a more flexible hierarchical transform structure comparing with RAHT and has the potential to compress larger point clouds effectively. In the near future, we would like to explore other geometry structures of the point cloud and better graph designs under the SSL framework.

\bibliographystyle{IEEE}
\bibliography{refs}
\end{document}